# Reliable Benchmarking of Artifact Detection in Computational Pathology: A Reproducibility and Uncertainty Analysis

Konstantinos Moutselos[1,*], Ilias Maglogiannis[1]

[1] Department of Digital Systems, University of Piraeus, Piraeus, Greece

* Corresponding author: kmouts@unipi.gr

ORCID — Konstantinos Moutselos: 0000-0002-6759-8540; Ilias Maglogiannis: 0000-0003-2860-399X

Email addresses: kmouts@unipi.gr (K. Moutselos), imaglo@unipi.gr (I. Maglogiannis)

## Abstract

**Background and Objective:** Automated quality control is a prerequisite for whole-slide image analysis, yet the benchmarks on which quality-control methods are compared share four properties that make their reported differences hard to interpret: few independent slides, annotation concentrated in a minority of them, pooled ratio metrics with no closed-form standard error, and a single inherited train/test partition. We propose a reliability protocol for such benchmarks.

**Methods:** The protocol quantifies four sources of variability — test-set sampling (stratified paired bootstrap with permutation tests), training stochasticity (a repeated training run), partition composition (a survey of random partitions), and undocumented preprocessing (decomposition of steps the published description omits) — and treats a claim as reportable only if it survives all four. Three of the four cost minutes of processor time; the fourth costs one training run. We apply it to an independent reconstruction of a published diffusion-based artifact detector, with every documented ambiguity resolved by explicit ablation, evaluated on the original 24-slide partition and against a supervised baseline.

**Results:** The method's central mechanism reproduces: the auxiliary contrastive term improves pooled F1 from 0.673 to 0.688 and replicates under a second seed (+0.0156, p = 0.031; +0.0190, p = 0.005), although it acts on pen marking rather than the artifact types cited to motivate it. Its comparative claims do not: differences between design variants, and against the supervised baseline, fall inside the uncertainty of the evaluation. Four of 24 slides carry 70% of scored annotated pixels, giving an effective sample size of 6.2, and the partition in general use sits at the 7th percentile. An unreported tissue-restriction step excludes 41.4% of out-of-focus annotation against 2.6% of air bubble; such a gate is confounded with blur by construction, and no threshold within the same family of criteria removes the bias.

**Conclusions:** Small-cohort quality-control benchmarks support far weaker conclusions than current reporting practice implies. The four checks proposed here are cheap enough to accompany any evaluation on such a resource; they separate reproducible effects from differences the evaluation cannot resolve, and detect structurally confounded preprocessing before any measurement is made.

## 1. Introduction

Whole-slide images carry artifacts that no downstream model was trained to expect: regions out of focus, tissue folded during sectioning, air trapped under the coverslip, ink left by a pathologist's pen. These are not rare. They arise from ordinary variation in a manual, physical, multi-step laboratory process [1], and they propagate silently — a tile of blurred tissue enters a multiple-instance-learning bag indistinguishably from a diagnostic one, and the slide-level models that consume such bags [2,3], increasingly trained under weak or few-shot supervision [4] where every tile carries weight, have no way to signal that anything is wrong [5–7].

Automated quality control is therefore a precondition for the reliability of everything built on top of it, and interest in it has grown accordingly.

What has not grown with it is the evidential standard applied to the resulting comparisons. Artifact detectors are typically reported as a table of point estimates on a single fixed partition of a small annotated cohort: one pooled F1 per configuration, one per-type sensitivity column, a comparison against one or two prior methods. Differences of one or two points in the second decimal place are presented as results. Rarely is there an interval, a repeated training run, or an indication of how much of the reported difference would survive a different draw of slides. The convention is not universal in the wider field — work on out-of-distribution detection in digital pathology has reported bootstrap confidence bounds throughout, and variability under repeated training has been studied in medical image segmentation — but it has not reached the artifact-detection literature. This is understandable. Exhaustively annotated whole-slide data is expensive, cohorts are correspondingly small, and the reporting convention is inherited from adjacent fields whose test sets contain thousands of independent samples. It is nonetheless a problem, and a specific one: the pixel-pooled metrics used here behave very differently from what that convention assumes.

Four properties make such a benchmark liable to produce conclusions that are precise in appearance and unsupported in fact. The unit of independence is the slide rather than the pixel, so a cohort of a few dozen slides supplies a few dozen observations however many millions of pixels it contains. Annotated area is distributed unevenly across those slides, so a minority of them governs the variance of every pooled statistic. The metrics themselves are pooled ratios rather than means, so they have no closed-form standard error and the familiar tests do not apply to them. And the train/test partition is drawn once and inherited, so every reported number is conditional on a draw whose representativeness is never examined. Each of these is individually well understood. Their conjunction, which is the normal condition of whole-slide quality control, is what makes the reported second decimal place uninterpretable.

We treat this as the object of study, and use diffusion-based pathology quality control as the case through which to examine it. The case is well chosen for the purpose rather than incidental to it: the method is recent, the benchmark is public and widely used, no reference implementation exists, and the published description — like most — leaves several decisions to the reader. Reconstructing it therefore exercises every source of variability at once, and does so on a resource that other groups are actively building on. Independent reproduction is in any case scarce in this field [8,9], for reasons that have been documented and that Section 1.2 returns to.

Our contribution is a protocol and an account of what it finds. The protocol quantifies four independent sources of variability — test-set sampling, training stochasticity, partition composition, and preprocessing steps that the published description omits — and requires a claim to survive all four before it is treated as reportable. Three of the four cost minutes of processor time; the fourth costs a single additional training run. All are cheap relative to the training they accompany, and none requires data or code beyond what a reproduction already has in hand.

Applied to the case study, the protocol separates the findings cleanly, and the separation is what we wish to report.

**The mechanism reproduces.** The auxiliary contrastive term that the method adds does improve detection; the improvement replicates across independently trained models, and it is not an artefact of test-set sampling. It is small, and it acts on a different artifact type than the one the original work invokes to motivate it, but it survives every axis.

**The comparisons do not.** Differences between the design variants, and between this method and a supervised baseline, all fall inside the uncertainty of the evaluation — including an ordering against that baseline which reverses in our hands without either direction being statistically supported. This is not a finding about the method. With ground truth concentrated in a handful of slides, the effective sample size of the 24-slide test partition is approximately six, computed over the ground truth actually scored, and the interval on any single pooled F1 spans nearly 0.4. A number that unstable is not a property of a method: repeat the study on a different draw of slides and it moves by more than any difference this literature reports.

**And one undocumented step is not merely undocumented but structurally confounded.** Every implementation must restrict predictions to tissue; no published description says how. The restriction we implemented excludes 41% of out-of-focus annotation against 3% of air bubble, shifting a headline per-type sensitivity by 0.36. The instructive part is not the size of the shift but the reason the obvious correction fails. Cheap tissue criteria threshold image properties — chromatic saturation, luminance, local gradient — that out-of-focus tissue is defined by degrading, so the gate is confounded with the artifact by construction, and no choice of threshold or disjunction of such properties separates them. An unspecified step can therefore leave a reconstruction with no correct choice available inside the space of choices it appears to offer.

Our contributions are: (i) a four-axis protocol for establishing which conclusions on a small, imbalanced benchmark survive evaluation uncertainty, together with an explicit decision rule and the reporting requirements that follow from it; (ii) a statement of the conditions under which such a protocol is warranted, so that it can be applied without reference to the case examined here; (iii) the finding that a preprocessing gate may be confounded with the phenomenon under study by construction, which is detectable before any measurement and is not remediable by adjusting the gate; (iv) an independent reconstruction of a published quality-control method, with every documented ambiguity resolved by explicit ablation rather than by selection, evaluated under all four axes and against a supervised baseline on a common denominator; and (v) a quantitative characterisation of a widely used artifact benchmark — its effective sample size, the position of its standard partition among possible partitions, the non-overlapping size distributions of its artifact classes, and one published per-type result the reconstruction does not recover — each of which is inherited by any work using the same resources.

The protocol is demonstrated on one method and one benchmark, and we do not claim empirical generality beyond that. What we do claim is that its four conditions are stated independently of the case, that they are satisfied by most public whole-slide quality-control resources, and that each of its four measurements is cheap enough to be performed as a matter of course.

### 1.1 Related work on artifact detection

Approaches divide broadly by supervision. Supervised segmentation models learn artifacts directly from exhaustive pixel annotation and perform strongly where such annotation exists; GrandQC, trained on 420 densely annotated slides [10], is the reference example and serves as our baseline. It pairs a learned tissue-detection module (Dice 0.957) with a seven-class artifact segmenter, and its authors released quality-control masks for the whole of TCGA, which is what we use here [11]. Earlier open tools take the same supervised route with less capacity: HistoQC [12] scores slides on handcrafted image metrics, while PathProfiler [13] and HistoROI [14] classify patches rather than segmenting pixels, and further work has addressed multi-class artifact segmentation and rapid tissue-versus-artifact separation [15,16]. Both the general question of whole-slide image quality and the specific handling of artifacts have been reviewed at length [17,18], and more recent pipelines combine handcrafted, deep and foundation-model features for the same task [19,20]. The common cost of the supervised route is the annotation itself, and its coverage is limited to the artifact categories present in the training set.

The alternative treats artifacts as out-of-distribution with respect to clean tissue. A generative model is fitted to clean tissue only, and anything it reconstructs poorly is flagged. Diffusion models are well suited to this, and the approach has an established line in medical imaging: partial diffusion followed by denoising, with the reconstruction error as an anomaly score, was developed as a general out-of-distribution detector [21] and applied to brain MRI [22], and has since been taken to whole-slide scale in digital pathology, where it has been benchmarked systematically against autoencoder- and GAN-based alternatives [23]. This requires no artifact annotation for training, and in principle detects artifact types never seen — an attractive property in a domain where the catalogue of possible failures is open-ended. The cost is that "poorly reconstructed" and "artifact" are not the same predicate, and the gap between them is where such methods succeed or fail.

### 1.2 Evaluation and uncertainty practice in biomedical image analysis

Two bodies of work bear on the reliability of the comparisons examined here, and they address different quantities. The vocabulary is shared, so the distinction is worth drawing: it is what locates our contribution.

**Uncertainty as an output of the model.** A large literature equips predictive models with a measure of their own confidence — Bayesian approximations, Monte Carlo dropout, deep ensembles, evidential formulations — and a review spanning 2013 to 2023 [24] shows the field to be mature. Within medical imaging the measure is per-voxel or per-structure and is used operationally: to flag segmentations likely to be poor, to weight a fusion of modalities, to drive semi-supervised training. Quality control appears in that literature in exactly this sense, as a map identifying outputs a clinician should re-examine. Calibration work asks the sharper question of whether a stated confidence matches an observed frequency, and scalable calibration of ensembles for segmentation [25] is the closest existing analogue to what we do, since it concerns the trustworthiness of a reported quantity rather than the accuracy of a prediction.

**Uncertainty of the reported measurement.** The quantity at issue here is different. It attaches to the number in the results table rather than to any prediction, and asks how far that number would move under circumstances fixed once and never varied: a different draw of test cases, a different training seed, a different partition, a different reading of a step the method description leaves open. A model can be perfectly calibrated and the comparison between it and a competitor still be unresolvable, because the resolution of the comparison belongs to the evaluation design rather than to either model. As far as we are aware this question has not been posed for whole-slide quality control, and only sporadically in adjacent areas — bootstrap bounds accompany every result in one benchmark of out-of-distribution detection in digital pathology [26], variability under repeated training has been examined in medical image segmentation [27], and the variance introduced by relying on a single held-out test set has been quantified through nested cross-validation in medical imaging more broadly [28] — but none of these practices is standard.

**Why pooled metrics make the distinction consequential.** Segmentation and detection results in this literature are ordinarily reported as pooled ratios: counts of true positives, false positives and false negatives are summed over all cases and the ratio formed once. Such a statistic is not a mean of per-case values, and treating it as one is the most common statistical error in this setting. It has no closed-form standard error; the tests that assume independent, identically distributed observations do not apply; and its variance is governed by whichever cases contribute most to the sums, which on an unevenly annotated cohort may be a small minority. The choice of metric is not neutral even before this — a comparative study of segmentation metrics on prostate MRI [29] shows that different measures of the same agreement rank the same methods differently and respond differently to object size and boundary definition. Our concern is one level further out: not which metric to report, but how much the reported value of any of them would move if the evaluation were repeated.

**Reproduction as the setting in which these questions become visible.** Independent reproduction is scarce in computational pathology, and documented as such: an audit of algorithms published between 2019 and 2021 found that 41 of 160 studies released code [8], 20 released trained weights, and 16 evaluated on an independent cohort, and a review of whole-slide image analysis reached similar conclusions [9]. The obstacles are themselves well described, and are structural rather than a matter of individual practice [30]. Scarcity matters here for a specific reason beyond the general one. A reconstruction must supply every step the source leaves unstated, and in doing so it enumerates the degrees of freedom that a single-implementation study never has to name. Preprocessing is where these accumulate. Stain normalisation, to take the best-studied case, is a routine step whose treatment varies widely between implementations and which materially alters the image statistics that later stages depend on; both the variability of its treatment across settings and its downstream consequences have been examined in this journal [24,31]. Tissue detection, patch selection and thresholding are less studied and no less consequential, and they are ordinarily reported in a sentence or not at all.

**Methods built on generative priors, and what they inherit.** The method examined in this paper detects artifacts as out-of-distribution deviations from a generative model of clean tissue. That design choice removes the need for artifact annotation during training but does not remove any of the above: the model still requires a tissue criterion, a patch selection rule and a decision threshold, and the benchmark it is scored on still has few slides and uneven annotation. Diffusion models have been applied in histopathology for synthesis, normalization [32] and anomaly detection, and the pathology foundation models now used as priors are trained on corpora [33] orders of magnitude larger than any quality-control benchmark. The asymmetry is worth stating plainly, because it recurs throughout what follows: the models have grown by several orders of magnitude while the resources used to compare them have not. A benchmark of two dozen slides is being asked to discriminate between methods trained on tens of thousands.

### 1.3 The method and benchmark examined

The method we reproduce takes the second route [34], using a pathology foundation model as the generative prior [33]. Clean patches are noised to a fixed diffusion timestep and denoised [35,36]; the per-pixel reconstruction error forms a heatmap in which artifacts appear as regions of elevated error. A low-rank adaptation (LoRA) [37] adapter is fine-tuned on clean tissue to sharpen the clean distribution.

To this the authors add an auxiliary contrastive objective. A small projection head, denoted `f_A`, maps latent representations into a space in which clean and artifact patches are pushed apart, on the argument that artifacts most similar in appearance to clean tissue — the paper names tissue folding and air bubbles — are otherwise reconstructed too well to be detected by error magnitude alone. Training minimises the reconstruction loss with this contrastive term added, weighted by a coefficient $\lambda$.

Three implementation details are not fully specified, and each admits more than one reading: whether the contrastive term operates on the output of `f_A` or on the raw latent; whether artifact patches for the contrastive set are sampled uniformly or balanced across types; and what minimum annotation overlap qualifies a patch as an artifact patch. Rather than choose, we trained all readings — the resulting configurations are described in Section 2.2. We note this not as criticism: incomplete specification is the norm, and it is precisely what makes reproduction informative.

The published evaluation reports a separation gap between clean and artifact latent representations as a training diagnostic, and pixel-level F1 with per-type sensitivity on the test partition, alongside a comparison with GrandQC.

The benchmark on which all of this is measured is equally specific. AIRAQC [38] provides artifact annotations on 50 whole-slide images from The Cancer Genome Atlas (TCGA) [39] across several cancer types, released as the test set of a multi-magnification supervised detector. It defines eight artifact classes and evaluates four of them — out-of-focus, pen marking, tissue folding and air bubble — and the method under study adopts the same four. Annotation is at whole-slide scale but is not exhaustive in the sense of GrandQC's training data — the original work itself notes that pen marking outside tissue is frequently unmarked [34], so precision is systematically underestimated for all methods evaluated on it.

The partition used by the original work, which we adopt, assigns 16 slides to training and 24 to test, out of 42 usable cases. Note that these 50 slides constitute AIRAQC's *entire* test set, 25 with artifacts and 25 without; the method under study therefore partitions an evaluation set into training and test material. That a QC method is evaluated on more slides than it is trained on is itself unusual, and it shapes much of what follows.

## 2. Materials and methods

### 2.1 A reliability protocol for small-cohort benchmarks

The protocol is stated here independently of the case study, so that it can be applied to any benchmark with the properties set out in Section 2.1.1. Each of its four components is cheap relative to the training it accompanies. Sections 2.2 to 2.7 instantiate it on the method and benchmark examined here, and Section 3 reports what it found.

### *2.1.1 When a benchmark needs this*

Four properties, jointly, make a benchmark liable to produce apparently precise but statistically unsupported conclusions. They are common in biomedical image analysis and near-universal in whole-slide quality control.

**Few independent units.** The unit of independence is the slide, the patient or the acquisition, not the pixel or the patch. A benchmark of a few dozen slides supplies a few dozen independent observations however many millions of pixels it contains, and reporting at pixel level obscures this by two or three orders of magnitude.

**Concentration of the target signal.** Annotated area is rarely distributed evenly across units. When a small minority of units carries most of the annotation, the variance of any pooled statistic is governed by that minority, and the nominal unit count overstates the information available. The Kish effective sample size [40] — the number of equally weighted units that would yield the same variance as the unequally weighted set actually held — makes this concrete and costs one line of arithmetic over the per-unit annotation counts.

**Pooled ratio metrics.** Pixel-level F1, Dice, sensitivity and precision are normally reported pooled: counts are summed across units and the ratio taken once. A pooled ratio is not a mean of per-unit values. It has no closed-form standard error, and the tests that assume a mean of independent observations — the t-test above all — do not apply to it. This is the point at which practice most often goes wrong, because the metric looks like an average and is treated as one.

**A single fixed partition.** Where the train/test split is inherited from a prior publication or drawn once and frozen, every reported number is conditional on that draw, and nothing in the reported result indicates how conditional.

Where all four hold and the methods under comparison differ by small margins — the second decimal place of a pooled metric is typical — the protocol below is warranted. Where the test set is large and the target signal evenly spread, it is unnecessary.

### *2.1.2 Axis 1 — Test-set sampling*

*If a different sample of units had been drawn from the same population, how different would the result be?*

Because the estimator is a pooled ratio, the interval must be obtained by resampling. Units are resampled with replacement, stratified on whatever property drives the variance — in artifact detection, the presence or absence of the target in the unit — and the pooled statistic is recomputed on each replicate, with the interval taken as percentiles of the resulting distribution [41,42]. Where two configurations are being compared and per-unit counts are held for both, the resampling must be **paired**: the same resampled units enter both sides of every replicate. Significance for a paired difference follows from a permutation test over unit-level sign flips rather than from the interval.

The distinction between levels and differences is the operative one. On a concentrated benchmark the interval on a level can be an order of magnitude wider than the interval on a paired difference between two closely related models, because the pairing removes the between-unit variance that dominates the former. Two consequences follow. First, differences between configurations evaluated in the same study remain resolvable even when the levels are not. Second, a comparison against a figure published elsewhere cannot be paired, since the per-unit counts behind it are not available, and no interval computed on one's own estimate licenses a claim about agreement with it.

One point of interpretation is easy to get wrong and worth stating explicitly. A resampling interval describes how much *one's own* estimate would move under a different draw of units. Where a published figure was measured on the *same* units, the difference between the two measurements contains no sampling error at all: it is exact, and it is a difference of implementation. Checking whether the published point estimate falls inside one's own interval tests nothing. What the interval does establish is that the estimator is unstable, and therefore that an absolute value on the benchmark is not a stable property of a method and should not be carried between papers.

**Cost:** minutes of CPU time, given saved per-unit counts.

#### *2.1.3 Axis 2 — Training stochasticity*

*If the same configuration were trained again, how different would the result be?*

At minimum, one configuration is retrained under a second seed and evaluated identically. Two runs do not support an interval, but they support the comparison that matters: whether a claimed effect is larger than the noise of retraining [27,43]. For two draws from a normal distribution the expected absolute difference is $E|X - Y| = 2\sigma/\sqrt{\pi} \approx 1.13\sigma$, so the observed difference $\Delta$ gives $\sigma \approx \Delta/1.13$. The estimate is unbiased but very imprecise, resting on a single pair, and should be used as an order-of-magnitude comparator rather than as the basis for intervals. Where budget allows more runs it should be spent here before it is spent on Axis 3.

The failure this catches is specific and is not caught by Axis 1: an effect can be stable under resampling of the test set and still be smaller than the difference between two independently trained instances of the same configuration. Such an effect is a property of one training run, not of the design choice it is attributed to.

**Cost:** one additional training run.

#### *2.1.4 Axis 3 — Partition composition*

*If a different subset of units had been assigned to training, how different would the result be?*

Answering this by retraining is expensive and, on small cohorts, usually unnecessary. The cheap precursor is a composition survey: draw a few hundred random partitions of the eligible pool at the same proportions as the real one, and compute for each the annotated area and unit count of every target class on both sides, together with the effective sample size of the resulting test set. Locating the real partition within these distributions as a percentile takes minutes and no training at all.

The survey serves as a gate on the expensive experiment and as a diagnostic in its own right. If training composition is stable across draws — no meaningful risk of a class being absent or severely under-represented in training — then retraining across partitions is unlikely to be informative, and the survey has saved the cost of finding that out. If the *test* side varies substantially, then part of the uncertainty reported under Axis 1 is a property of the particular draw rather than an inherent limit of the cohort, and should be reported as such. A partition at an unfavourable percentile of effective sample size will yield wider intervals than the cohort itself requires, which is worth knowing before that width is attributed to the benchmark.

**Cost:** minutes of CPU time on the annotation statistics alone.

#### *2.1.5 Axis 4 — Undocumented preprocessing*

*If a step the published description omits had been implemented differently, how different would the result be?*

Every method description leaves some operations unstated that any working implementation must nonetheless perform. These are not marginal: they are omitted precisely because they are regarded as routine, and routine steps can interact with the phenomenon being measured. The procedure has three parts.

First, enumerate the steps the reconstruction had to supply that the source does not specify — the test is whether a competent reader could produce them from the text alone, not whether they are conceptually obvious. Second, implement at least two defensible variants of each and report the spread in the affected metrics. Third, and most importantly, check whether that spread is **uniform across subgroups**. A preprocessing step can be close to neutral in aggregate while shifting per-class figures substantially, if it removes evidence of one class more readily than another. Where such a step excludes ground truth unevenly, the excluded material is typically never charged as a false negative, so per-class results computed after it are not on a common denominator and any conclusion of the form *this method handles class X better than class Y* inherits the distortion. Comparisons *within* a class are unaffected provided the same step is applied to every method.

A prior check precedes all three parts and costs nothing: name the image property each unspecified gate operates on, and ask whether the phenomenon under study alters that property. Where it does, the gate is confounded by construction, non-uniform exclusion should be expected rather than discovered, and widening the gate within the same family of criteria will not remedy it.

This axis differs from the first three in kind. They quantify variability under resampling; this one quantifies the consequence of an interpretive choice, and its output is a range attributable to reconstruction rather than to chance. It is also the cheapest of the four, requiring no retraining — only re-evaluation under the alternative variant.

**Cost:** re-evaluation only.

#### *2.1.6 Decision rule and reporting*

A quantitative claim is treated as reportable only if it survives all four axes: it clears the paired resampling interval, it exceeds the retraining noise, it is not an artefact of the partition in use, and it does not depend on an unspecified implementation choice. A claim that clears some but not others should be reported with the axis it fails named, not withheld and not asserted.

The corresponding reporting requirements are five, and none of them requires additional training:

1. **Report paired differences with intervals rather than levels.** Where levels are reported, they carry their own interval and an explicit statement that they are not comparable across studies. 2. **Report the effective sample size** alongside the unit count. A unit count on a concentrated benchmark communicates the opposite of what it appears to. 3. **Report the seed** **repeat**, including when it changes nothing — a null result on this axis is informative and is currently almost never available. 4. **Report per-class results with their coverage denominator**, or remove the unevenness that makes the denominators differ. 5. **Report the partition percentile**, so that readers can distinguish a limitation of the cohort from a limitation of the split.

**Table 1.** The protocol in summary. Each axis answers a different counterfactual, and a quantitative claim is treated as reportable only if it survives all four.

| Axis | Counterfactual | Procedure | Cost | Failure it catches |
|---|---|---|---|---|
| Test-set sampling | A different draw of units | Stratified paired bootstrap; permutation test on unit-level sign flips | Minutes CPU | A difference within the noise of which units were sampled |
| Training stochasticity | A second training run | Retrain ≥1 configuration under a new seed; $\hat{\sigma} \approx \Delta/1.13$ | One training run | An effect smaller than the variability of retraining |
| Partition composition | A different train/test split | Survey a few hundred random partitions; Kish effective n; locate the real split as a percentile | Minutes CPU | Uncertainty attributed to the cohort that belongs to the draw |
| Undocumented preprocessing | A different reading of an unstated step | Implement ≥2 defensible variants; report spread and its uniformity across subgroups | Re-evaluation only | A per-class conclusion that depends on an unreported choice |

### 2.2 Reconstruction of the method

We reconstructed the method from its published description; no reference implementation was available. A pathology foundation model (PixCell-1024) is used as a diffusion backbone [33]. Clean tissue patches are noised to a fixed timestep and denoised, and the per-pixel reconstruction error serves as an out-of-distribution score: tissue resembling the clean training distribution reconstructs well, artifacts do not. A LoRA adapter is fine-tuned on clean patches, optionally with an auxiliary contrastive term that pushes apart the latent representations of clean and artifact patches.

Where the description was ambiguous we enumerated the plausible readings and trained each. Three points required a decision. First, the contrastive term can be applied either to the projected features of the auxiliary head `f_A` or to the raw latent `z` that the encoder produces before the head; we trained both (variant A and variant B). Second, artifact patches for the contrastive set can be sampled uniformly or balanced by artifact type; we trained both. Third, the minimum annotation overlap for a patch to be considered an artifact patch is unspecified; we trained at 0.05 and 0.5. Together with a baseline at $\lambda = 0$ — no contrastive term at all — this gives five models (configuration grid in Supplement S5). All share the same data pipeline, patch catalogue and inference procedure, so differences between them are attributable to the choice under test.

Patch extraction is where our reconstruction diverges most from the published account, by enough to record here rather than in a footnote. The source reports 63,000 clean patches and 400 artifact patches for training. Ours yields **2,185 clean patches** from our same 16 slides — a factor of twenty-nine fewer — alongside 4,965 candidate artifact patches at a minimum annotation overlap of 0.05, from which 400 are sampled as described, and 3,228 at an overlap of 0.5. The parameters that would account for the difference are not specified: the stride between patches, the minimum tissue fraction for a patch to count as clean, and the magnification at which the patch size is defined. We use a stride of 512 pixels and require half of a patch to be tissue; a shorter stride alone would multiply the count several-fold. This matters because a model that has seen 2,185 distinct clean patches has a poorer picture of the clean distribution than one that has seen 63,000, and Section 3.3 returns to it.

Whole-slide inference produces a reconstruction-error heatmap at 10× on a strided grid. Post-processing follows the published description [34]: Gaussian smoothing, clipping to a user-defined range [v_min, v_max], a per-slide Otsu threshold [44] within that range, and morphological closing followed by opening. Two additions are ours and are flagged as such throughout. First, v_min and v_max are described as user-defined without a stated procedure, so we select them per model by grid search maximising pooled F1 — optimistic, since each reported figure is then a maximum over 25 grid points evaluated on the test set itself. Second, the published description specifies **no restriction of predictions to tissue**; we impose one, without which false positives on background dominate. Section 2.4 states what it is and Section 3.4 measures what it costs, because it turns out not to be neutral between artifact types. Note also that the Otsu threshold is computed over pixels inside that restriction, so the addition affects the threshold itself and not only which pixels are scored.

### 2.3 Data and ground truth

We use the AIRAQC annotations on haematoxylin and eosin (H&E) stained TCGA whole-slide images, following the original partition: 16 training and 24 test slides drawn from 42 eligible cases, with 8 excluded as unusable or insufficiently tissue-bearing. Two further slides serve as a validation set. The partition is worth stating carefully, because the source describes it three different ways: its dataset section gives 16 training and 24 test slides, its results table lists 24 training slides for its own models, and its discussion says 14. We follow the dataset section. Sixteen plus twenty-four is also forty, against a stated cohort of fifty, and the remaining ten are not accounted for. Two slides that appear in the original publication's figures are pinned to the test set so that our qualitative results can be compared against theirs (Fig. 2).

Ground truth is decoded from the annotation images by nearest palette colour, subject to a maximum Euclidean distance of 60 in RGB; pixels beyond that distance — anti-aliased boundaries, and colours outside the documented palette — are excluded from per-type masks rather than force-assigned to the nearest entry. AIRAQC detects eight artifact classes but evaluates four — out-of-focus, pen marking, tissue folding and air bubble — and the method under study adopts the same four. We restrict to those four. Two observations on the released annotations are worth recording for anyone reproducing this work: a small fraction of annotated pixels (0.3%) are alpha-blended boundary values lying far from every palette colour, which a nearest-colour decoder without a distance cutoff will silently assign to whichever class sits closest to them; and 24,440 pixels carry an exact colour outside the four evaluated classes, presumably one of the four AIRAQC defines but does not evaluate. Both are excluded from per-type masks by the distance cutoff. The supervised baseline in Section 3.3 emits a "dark spot and foreign object" channel with no counterpart among the four evaluated types, which affects how it must be scored.

All metrics are computed at pixel level inside the refined coverage mask, and pooled. F1 is the harmonic mean of precision and sensitivity, equivalently the Dice coefficient between the predicted and annotated artifact areas; sensitivity is the fraction of annotated artifact pixels recovered, precision the fraction of predicted artifact pixels that are annotated. Pooling means that true positives, false positives and false negatives are summed across slides and the ratio taken once. Pooled F1 is therefore not a mean of per-slide F1 values, a distinction that determines the statistical treatment below.

**2.4 Instantiating the protocol**

Section 2.1 states the protocol in general terms. This subsection records the parameter choices that instantiate it on the present benchmark and, where a choice was forced by a property of this data, says which property. Fig. 1 summarises the pipeline, the points at which the published description leaves a decision open, and the four axes below.

The benchmark satisfies all four conditions of Section 2.1.1. The unit of independence is the slide and there are 24 of them in the test partition; annotated area is heavily concentrated across those slides, as Section 3.4 quantifies; every reported metric is a pooled ratio; and the partition is inherited from the original work rather than drawn afresh. The configurations under comparison differ in the second decimal place of pooled F1.

**Axis 1 — Test-set sampling.** Each of 20,000 replicates draws slides with replacement, sums their counts and recomputes F1 once, reproducing the estimator exactly. Two design choices are forced by the data. Resampling is **paired**, using the same resampled slide indices for every model within a replicate: per-slide F1 ranges from 0.041 to 0.954 here (per-slide values in Supplement S1), so slide difficulty dominates every other source of variation and independent draws per model would compare which model received the easy slides. It is also **stratified**, preserving 13 artifact and 11 clean slides, because unstratified replicates occasionally contain no clean slides at all and leave the clean-slide false-positive rate undefined; the 13/11 composition is a fixed property of the test set, not a quantity being estimated. We report percentile intervals alongside a paired permutation test on pooled F1 over 100,000 sign-flips across slides, as a distribution-free check.

**Axis 2 — Training stochasticity.** We retrained variant A under a second seed, which changes LoRA initialisation, the artifact patches sampled for the contrastive set, batch ordering and noise draws — total run-to-run variance, which is the quantity the axis requires. Both resulting checkpoints were evaluated: one matched to the first run's training step, for a comparison in which only the seed differs, and one at its own selection criterion, kept as a check that step-matching does not understate the run — it scores 0.695 against the matched checkpoint's 0.692, and enters no comparison. Every seed-7 figure reported in this paper — the interval, the p-value and the seed-to-seed difference — comes from the matched-step checkpoint, because the axis requires that only the seed vary. Both were scored with the **first run's** post-processing thresholds, so that no part of the observed difference is attributable to re-tuning. The $\hat{\sigma} \approx \Delta/1.13$ estimator of Section 2.1.3 is applied to the matched-step pair and used only as an order-of-magnitude comparator, as that subsection requires.

**Axis 3 — Partition composition.** Retraining under a different partition costs approximately 6 GPU-hours, so the survey precedes it as a gate. We drew 200 random partitions from the 42 eligible slides at the same 24/16/2 proportions as the real one and computed, for each, the annotated pixels and slide counts of every artifact type on both sides, together with the Kish effective sample size of the resulting test set. The real partition is then located within each distribution as a percentile. The survey measures composition, not performance; whether the expensive experiment was in the end warranted is reported in Section 3.4.

**Axis 4 — Undocumented preprocessing.** The published description specifies its post-processing completely — Gaussian smoothing, clipping, per-slide Otsu thresholding, morphological closing then opening — with one exception: it says nothing about restricting predictions to tissue. Some restriction is unavoidable, since without one false positives on background dominate every metric, so this is a step any implementation must supply and no reader could derive from the text: precisely the criterion of Section 2.1.5.

Ours is the intersection of two components, and the axis requires both to be named. The first is the footprint of the inference grid: patches are run only where the tissue fraction of the patch reaches `tissue_threshold`, set to 0.1, and the error map is defined only where patches were run. That parameter is a third undocumented degree of freedom, and a tuned one — a more permissive value measurably reduced precision in earlier testing — although Section 3.4 shows it nearly inert with respect to the ground-truth denominator. The second is a pixel-level tissue mask computed on the slide thumbnail as *saturation above its Otsu threshold, or value below its own*, upsampled to the error-map resolution. Two features of the composite matter here. It acts at **two** points rather than one: it determines which pixels are scored and, because the Otsu threshold is computed over the pixels inside it, the decision threshold as well. And its effect on the ground-truth denominator decomposes exactly, by intersecting each annotated type mask with each component separately, which localises the exclusion to one component of the composite or to both jointly. Section 3.4 reports the decomposition and Section 4.2 discusses it. The instantiation is nonetheless weaker than Section 2.1.5 prescribes, and we mark it as such. A full treatment would re-run the pipeline under a second, differently constructed tissue criterion and report the spread in every metric, including those affected through the decision threshold. What we report is the spread attributable to the denominator alone, which is a lower bound on the total, together with a mechanistic account of why a second cheap criterion would not be expected to differ. Section 4.4 returns to this.

### 2.5 Component-level analysis

Per-slide F1 varies by more than twenty-fold across the test set, and per-type sensitivities differ by a factor of seven. Because artifact types differ systematically in size — out-of-focus regions are diffuse and large, folds are small and numerous — the two are confounded at slide level. We therefore analysed at the level of individual annotated connected components, labelling each type mask separately so that adjacent artifacts of different types are not merged, and restricting to the same refined coverage mask used for the metrics. Each component contributes one observation: its size, its type, and the fraction of it covered by the prediction. Agreement between types at equal size would indicate that type is a proxy for scale; separation would indicate that type has independent standing. The analysis also records the size distribution of each type's annotation, because whether the comparison is well posed at all depends on whether the types occupy overlapping ranges of size.

### 2.6 Supervised baseline

We compare against GrandQC [10], a supervised segmentation model trained on 420 exhaustively annotated slides. It emits seven classes: no artifact, air bubble/slide edge, pen marking, out-of-focus, tissue fold, dark spot/foreign object, and background — air bubble with slide edge, and dark spot with foreign object, being merged pairs in the published class scheme. We use the authors' **own released masks** [11] **for the entire TCGA cohort** rather than running inference ourselves, so no reimplementation risk attaches to the baseline. They were available for 21 of our 24 test slides; the three missing all carry frozen-section portion codes, and the pipeline appears not to have been applied to frozen sections. All three are clean slides, so every artifact slide is present and only the clean-slide false-positive rate loses power.

To make the comparison paired rather than merely juxtaposed, the baseline is scored **inside our refined coverage mask**, at our resolution, with our confusion counting. Scoring it under its own tissue mask would compare the two methods on different denominators. Its class labels were mapped to ours by measured overlap on three slides of differing composition rather than assumed from the label names; the resulting assignment matches the published class scheme. The mask release states its class order explicitly - tissue, fold, dark spot and foreign object, pen marker, edge and air bubble, out of focus, background - and our overlap-derived assignment reproduces it exactly, on all seven. The masks are distributed at 1.5 um/px, which we independently confirmed from the mask-to-slide dimension ratio (5.95 across three slides at 0.252 um/px). This matters for Section 3.3: of GrandQC's three published variants, 1.5 um/px corresponds to the 7x version, which is the one in which the dark spot / foreign object class is least accurate (Dice 0.325, against 0.406 at 5x and 0.528 at 10x) [10]. It also means the masks are finer than our error maps (8 um per pixel), so aligning them downsamples rather than interpolates. One label appears in the masks but not in the published class list - value 0, covering a few percent of some slides - which we treat as non-artifact, the conservative reading.

### 2.7 Verification

Every evaluation is re-derivable. Post-processing thresholds are stored with each result and can be replayed deterministically, so re-running an evaluation reproduces its published figures exactly rather than approximately. All analysis scripts assert that per-slide counts re-pool to the saved aggregate before proceeding, and abort otherwise; this check caught three separate errors during the study, including one in the component analysis that would have produced a plausible but wrong result.

## 3. Results

### 3.1 The mechanism reproduces

Adding the contrastive term with `f_A` inside the reconstruction loss improves pooled F1 from 0.6725 to 0.6881. The paired bootstrap places the difference at **+0.0156, 95% confidence interval (CI) [+0.0035, +0.0339], paired permutation p = 0.031** (Fig. 3). It is also significant on artifact-bearing slides alone (+0.0147, [+0.0037, +0.0327]), so it reflects improved segmentation rather than merely fewer false alarms on clean tissue.

The independently trained second seed **replicates** the finding, more strongly: **+0.0190, [+0.0072, +0.0390], p = 0.005**. The seed-to-seed difference in pooled F1 is 0.0035, so the effect is four to five times the variability of retraining the same configuration, as well as excluding zero under test-set variability.

The mechanism is specific. Per-type sensitivity gains over the baseline are confined to pen marking: **+0.0385 [+0.0227, +0.0646]** in the first seed and **+0.0325 [+0.0072, +0.0552]** in the second, both excluding zero, while out-of-focus and tissue folding move negligibly (+0.0008 [−0.0054, +0.0178] and +0.0046 [−0.0220, +0.0610], both crossing zero). Air bubble moves by +0.216 in the point estimate, but its interval spans nearly the unit interval and the difference does not exclude zero (Section 3.4), so we do not report it as a gain. This is notable because the original work motivates the contrastive term by the difficulty of tissue folding and air bubbles, which resemble clean tissue in appearance. The measurable benefit is elsewhere.

### 3.2 What does not survive

**Differences between the ambiguous design choices are not resolvable.** Applying the contrastive term to the raw latent instead of the projected features yields 0.6761; balanced artifact sampling 0.6727; a minimum overlap of 0.5 yields 0.6753. None differs significantly from the $\lambda = 0$ baseline ($p = 0.884, 0.993, 0.907$), and the difference between variants A and B is −0.0119 with CI [−0.0606, +0.0310], $p = 0.553$.

The reason is visible per type, and it is not that the variants are equivalent. Against variant A, the variant B family gains significantly on pen marking (+0.0804 [+0.0333, +0.1144]) and folding (+0.0391 [+0.0099, +0.0668]) while **losing** significantly on out-of-focus (−0.0407 [−0.0785, −0.0201]). Out-of-focus accounts for 53% of annotated pixels, so against the $\lambda = 0$ baseline a three-to-four point loss there cancels a twelve-point pen-marking gain in the pooled figure. The choice demonstrably matters; it simply does not matter for the summary metric. Sampling scheme and minimum overlap, by contrast, are genuinely inert — the three variant B runs are within noise of one another on every statistic.

**Operating-point claims do not survive seed variance.** Variant A produces fewer false positives on clean slides than variant B (+0.0199, bootstrap CI [+0.0084, +0.0460]), which under test-set uncertainty alone appears significant. But seed-to-seed variance on that same metric is 0.0133 — two-thirds of the effect. Two identical training runs, differing only in seed, produce clean-FP rates of 4.15% and 2.82%, a difference whose own confidence interval excludes zero ([−0.0245, −0.0062]), as do their out-of-focus (−0.0123 [−0.0397, −0.0045]) and air-bubble (−0.0123 [−0.0194, −0.0014]) sensitivities. Pooled F1 is stable across seeds while its components are not: the two runs occupy different points on the same precision–sensitivity curve. Seed selects an operating point, not a quality level.

This yields a reporting rule, which we state as our own heuristic rather than an established criterion: **a difference that excludes zero under a bootstrap but is smaller than roughly twice the seed-to-seed standard deviation should not be reported as a finding.** The factor of two borrows the familiar two-standard-deviation convention and is not calibrated; with $\sigma$ estimated from a single pair of runs it could not be. The principle behind it is not ours — benchmarking is known to involve several sources of variation [43] of which the random seed is only one, and holding them fixed rather than randomising over them understates uncertainty. What this study adds is that the two axes dominate different metrics: pooled F1 is governed by test-set sampling, while operating-point metrics such as the clean-slide false-positive rate are governed by the seed. Several comparisons in this literature, including some of our own initial ones, fall below that line.

### 3.3 The supervised baseline

GrandQC scores 0.755 on our slides under our tissue mask against its published 0.762, which validates the class mapping, resolution handling and restriction at once. Our best configuration scores 0.688 (0.692 in the second seed). The published comparison places the diffusion method ahead of GrandQC, 0.784 to 0.762; **our reproduction reverses the ordering**, and does so despite four respects in which the comparison favours us: thresholds are a grid maximum tuned on this test set, the baseline runs unadapted, it is scored inside our own mask, and that mask removes the off-tissue pen ink whose omission the original work states makes its precision an underestimate. Scored as the original describes, our shortfall would be larger.

**Neither ordering is statistically resolvable.** The paired difference is +0.0622 with CI **[−0.068, +0.213]**, p = 0.404. The interval is ten times wider than the one against our own baseline (±0.015), because pairing removes variance only insofar as two models agree slide by slide: GrandQC gains 0.38 on one slide and loses 0.24 on another. The same argument applies to the original claim — a 0.022 lead inside an interval of comparable width is not evidence of superiority in either direction.

The per-type comparison, by contrast, is one-sided. The baseline leads on all four types, three of them excluding zero: pen marking +0.454 [+0.268, +0.663], air bubble +0.391 [+0.050, +0.916], out-of-focus +0.071 [+0.024, +0.228]. **There is no artifact type on which the reproduced method leads.** The pooled near-tie decomposes into much better sensitivity (0.964 against 0.718) and worse precision (0.620 against 0.660). Pen marking is the pointed case: it is where the contrastive term produces its one reproducible gain of four points, and the baseline is forty-five ahead.

One scoring decision exceeds the entire difference between the methods. The baseline emits a "dark spot and foreign object" channel with no counterpart in the AIRAQC annotations. Including it gives 0.755; excluding it gives 0.691 (Fig. 2), a dead heat with our reconstruction ($\Delta = -0.0024$, $p = 0.983$). We take the inclusive figure as primary, because the task is binary and the reconstructed models do not distinguish types, so requiring type agreement of the baseline alone would impose a stricter criterion for the same task. Both must be reported, because choosing one is choosing the conclusion.

**Table 2.** All evaluated runs on the 24-slide test partition, four-type ground truth. Sensitivity, precision and pooled F1 are pixel-level over all slides; clean-FP is the false-positive rate on the 11 artifact-free slides. Per-type columns are sensitivity. GrandQC is scored inside the same coverage mask and on the 21 slides for which released masks exist. 95% intervals for every paired difference are in Fig. 3.

| run | sens | prec | F1 | OOF | pen | fold | air | clean-FP |
|---|---|---|---|---|---|---|---|---|
| basic (λ = 0) | 0.699 | 0.648 | **0.673** | 0.891 | 0.487 | 0.450 | 0.361 | 4.45% |
| variant A, seed 0 | 0.718 | 0.660 | **0.688** | 0.892 | 0.526 | 0.455 | 0.577 | 4.15% |
| variant A, seed 7 | 0.709 | 0.675 | **0.692** | 0.880 | 0.520 | 0.450 | 0.564 | 2.82% |
| variant B | 0.732 | 0.628 | **0.676** | 0.851 | 0.606 | 0.494 | 0.604 | 6.14% |
| variant B, balanced | 0.732 | 0.622 | **0.673** | 0.857 | 0.601 | 0.484 | 0.604 | 6.59% |
| variant B, overlap 0.5 | 0.736 | 0.624 | **0.675** | 0.861 | 0.605 | 0.482 | 0.604 | 6.74% |
| GrandQC | 0.964 | 0.620 | **0.755** | 0.963 | 0.980 | 0.819 | 0.967 | 4.04% |
| GrandQC, no class 3 | 0.836 | 0.588 | **0.691** | 0.724 | 0.978 | 0.813 | 0.967 | 3.99% |

**One published figure we do not reproduce.** Pen-marking sensitivity is reported at 0.866 for the enhanced model; our best reconstruction reaches 0.526 and no variant exceeds 0.606. The coverage mask accounts for at most 4.1% of pen-marking annotation (Section 3.4) and cannot close a gap of this size. The largest documented difference between the setups is the training

data — 2,185 clean patches against the 63,000 reported (Section 2.2) — but the extraction parameters are not published, so we cannot test this and do not claim it as the cause. It is the one place where the reconstruction clearly fails to recover a reported result, and we record it as such rather than absorbing it into an interval.

### 3.4 What the benchmark can support

The pattern above — mechanisms reproduce, comparisons do not — is a property of the evaluation, and it is characterisable. Fig. 4 shows the distribution that produces it.

**An unspecified implementation choice shifts one type's sensitivity by 0.36.** The tissue restriction described in Section 2.4 excludes 28.6% of all annotated pixels — but **41.4% of out-of-focus annotation against 2.6% of air bubble**. Excluded ground truth is never charged as a false negative, so the reported out-of-focus sensitivity of 0.892 is computed over 58.6% of the annotation; measured over all of it, the same predictions score 0.527.

Decomposing the exclusion locates it. The inference footprint accounts for 0.7 percentage points of the out-of-focus loss and the tissue mask for the remaining 40.7 — the patch threshold is therefore nearly inert with respect to the ground-truth denominator, and the dual tissue criterion is responsible. That criterion is a disjunction, and a disjunction normally widens coverage; here both of its branches fail on the same material and for the same reason. Blur reduces chromatic saturation and raises luminance together, so heavily blurred tissue falls below the saturation threshold *and* above the value threshold at once. Of the 40.7 points attributed to the mask, 18.9 lie on material the patch threshold had independently skipped (full decomposition in Supplement S3).

The exclusion tracks the severity of the blur rather than the class label. Within a single slide, where stain, scanner and both thresholds are held fixed, out-of-focus annotation on `TCGA-62-A46P` has a median saturation of 35 and is retained at 49.0%, while tissue folding on the same slide has a median of 112 and is retained at 88.5%. On the two test slides where out-of-focus annotation carries no saturation deficit relative to the other annotated types, it is retained at 88.8% and 88.4% — indistinguishable from folding. The pooled 41.4% is thus dominated by a small number of heavily blurred slides, one of which contributes 4.93 million of the 8.01 million annotated out-of-focus pixels, and is not a property of the class as such.

Within-type comparisons are unaffected, since the same mask applies to every method, and no pooled figure here depends on the choice beyond its being held constant. Comparisons *between* types are distorted, and any conclusion of the form *this method handles X better than Y* inherits the distortion.

**Our test partition is unusually unfavourable.** Across 200 random partitions of the eligible pool, training composition is stable — an artifact type is entirely absent from training in only 1.0% of partitions, and the real partition sits at percentiles 32–60 on training material for the three types that matter. The test side is another matter. Our test set has a Kish effective n of 3.88 against a median of 5.35 (**7th percentile**), its largest slide holds 45% of the ground truth against a median of 33% (**94th percentile**), and it contains 14 slides bearing any annotation against a median of 17 (**2nd percentile**; full distribution in Supplement S2). The wide intervals reported above are therefore partly an unlucky draw rather than solely an inherent limit of the benchmark: a typical partition would afford roughly 40% more effective sample size. Since training composition is stable, we did not retrain across partitions.

The two figures weight differently. Over annotated area the effective sample size is 3.88 and the largest slide holds 45% of the ground truth; over the ground truth actually scored, 6.15 and 27%. The difference is not a gain in information. The coverage restriction removes 41.4% of out-of-focus annotation, which lowers that type's share of the total from 71% to 54%, and because out-of-focus is also the most concentrated type the surviving weights are more even than the annotation warrants. The same step that distorts the per-type comparison makes the effective sample size look larger than the annotation supports.

A related imbalance is worth recording: the training partition carries forty times the air-bubble annotation of the test set (15.6M against 395k pixels). And what the evaluation can measure of it is small: 96k scored pixels spread over two slides, 61% of them in one, with an interval spanning [0.002, 0.950]. The models are heavily trained on a type that is effectively unmeasurable at evaluation.

**Table 3.** Published figures against this reproduction, on the same 24 slides. The interval in the last column describes how much *our* estimate would move under a different draw of slides; it is not a test of agreement with the published value, which was measured on these same slides. Differences here are differences of implementation, exact for this data, and we cannot assess whether they would persist elsewhere without the original per-slide counts. The intervals are given because they show the estimator to be unstable enough that no absolute figure on this benchmark should be read as a fixed property of a method.

| | **published** | **reproduced** | **95% CI** |
|---|---|---|---|
| basic, pooled F1 | 0.750 | 0.673 | [0.441, 0.835] |
| enhanced, pooled F1 | 0.784 | 0.688 / 0.692 | [0.468, 0.843] |
| GrandQC, pooled F1 | 0.762 | 0.755 | [0.618, 0.831] |
| enhanced, out-of-focus | 0.953 | 0.892 | [0.730, 0.936] |
| enhanced, pen marking | 0.866 | 0.526 | [0.314, 0.717] |
| enhanced, tissue folding | 0.723 | 0.455 | [0.069, 0.933] |
| enhanced, air bubble | 0.026 | 0.577 | [0.002, 0.950] |

### 3.5 Scale as a hidden denominator

**The artifact types are not measured on a common scale.** Analysed at the level of individual annotated connected components, detected fraction rises with component size within every type, and the types remain far apart at equal size (Fig. 5). Among components of 1,000–10,000 pixels the same model recovers 0.776 of out-of-focus, 0.538 of pen marking, 0.390 of folding and 0.000 of air bubble. The four appear to converge in the 10,000–100,000 band, to between 0.72 and 0.84, but that band contains two folding components and two air bubble components in the entire test set, and the apparent agreement should not be read as one. Above it the separation returns: in the 100,000–1,000,000 band out-of-focus scores 0.916 against pen marking's 0.486, and neither folding nor air bubble has a single component there.

The size distributions are the reason, and they barely overlap where the annotated area lies. Out-of-focus and pen marking have components reaching 1.32 million and 916,000 pixels respectively; the largest fold is 74,000 and the largest air bubble 58,000. Meanwhile the annotated area is overwhelmingly concentrated in the largest components: those above 100,000 pixels are 0.1% of all components and carry 80.8% of annotated pixels, while everything below 10,000 pixels — 99.6% of components by count — carries 7.1% (full component tables in Supplement S4). A pooled per-type sensitivity is a pixel-weighted average over this distribution. For out-of-focus and pen marking it is therefore a measurement made almost entirely on very large regions; for folding and air bubble it is a measurement made on small ones, because large ones do not exist. The two are not comparable quantities, and the difference between them is not a statement about how well the method handles each type.

This is the same failure as the coverage mask, in a different variable. There the types were scored against different denominators of annotated area; here they are scored over different ranges of object size. In both cases a per-type table presents as a like-for-like comparison something that is not one, and in both cases the distortion is a property of the benchmark that any method evaluated on it inherits.

## 4. Discussion

### 4.1 Implications for the method

The contrastive term works, and the evidence for it is now considerably stronger than in the original report: an effect of +0.016 to +0.019 in pooled F1 that excludes zero under a paired bootstrap, replicates under an independently trained seed, and is four to five times the difference between two runs of the same configuration. It is the one claim in this study that clears all four axes, and few reported effects in this literature have been tested that way.

Two qualifications matter for how it should be understood. First, the effect is small in a practical sense: two points of F1 on a metric whose own interval is nineteen points wide. It is a real improvement to a component, not a change in what the method can do. Second, and more interesting, it acts on pen marking, while the stated motivation is tissue folding and air bubbles. Pen ink is not subtle — it is saturated, high-contrast and unlike tissue in every respect — so the contrastive term appears to help where the out-of-distribution signal was already strongest, rather than where it was weakest. Whether this reflects the training composition, in which pen marking is abundant, or something about what a contrastive objective in latent space can actually separate, we cannot determine from this experiment. It is the most interesting question the reconstruction raises, and it needs a different experiment to answer.

That the design ambiguities turned out not to matter for the pooled metric is a useful negative result for anyone attempting the same reconstruction — with the caveat, which the per-type analysis makes clear, that "does not matter" holds for the summary figure and not for behaviour. Placing the contrastive term on the raw latent rather than on `f_A` significantly improves pen marking and folding and significantly degrades out-of-focus; these cancel because out-of-focus dominates the annotated area. A practitioner selecting a variant for a particular artifact profile would care a great deal about a choice that the benchmark's headline number reports as irrelevant. This is a general hazard of pooled metrics on compositionally skewed data, not a peculiarity of this method: a summary figure can be insensitive to a design choice precisely because the choice trades one subpopulation against another in the proportions the benchmark happens to hold.

### 4.2 Implications for benchmark design and reporting

The more transferable findings concern the evaluation resource rather than the method.

**Statistical power.** With four of 24 slides carrying 70% of the scored annotated pixels, the effective sample size over the ground truth actually scored is 6.2 and the interval on a single pooled F1 is roughly ±0.19. The consequence is not that published figures agree with ours — measured on the same slides, the differences between them are exact and reflect implementation, not chance. It is that a number of this instability should not be treated as a property of a method at all: repeat the study on a different draw of 24 slides and it moves by more than any difference reported in this literature. Paired differences remain informative, and remain sharp at ±0.015 for closely related models, which is why we report differences rather than levels throughout. The distinction is not pedantic. It changes which sentences a paper on this data is entitled to write.

**A type-dependent bias introduced by an unspecified step, and the failure of the obvious remedy.** The published method describes its post-processing completely except for one thing: it says nothing about restricting predictions to tissue. Some such restriction is unavoidable, and ours moves a headline per-type figure by 0.36 — two faithful reconstructions could differ by that much while both following the published description exactly. That much is the expected form of the problem. What is less expected is where the correction has to come from.

The natural reading of an uneven exclusion is that the criterion was poorly chosen and that a broader one would fix it. Ours already *is* the broader one — a disjunction of low saturation and high luminance, exactly the widening a saturation-only gate would call for — and it does not help, because the two branches are not independent tests: optical blur attenuates chromatic contrast and lifts mean luminance through the same physical mechanism, so the same tissue fails both conditions at once. Gradient- and texture-based criteria would fare worse, measuring precisely the high spatial frequencies blur removes.

The problem is therefore not a badly chosen threshold within an appropriate class of criteria; it is that the whole class of cheap appearance heuristics is confounded with this artifact by construction, because out-of-focus tissue is *defined* by degradation of the very properties such heuristics measure. Only a criterion trained to recognise blurred tissue as tissue — a learned segmentation decoder of the kind the benchmark's own annotators use [38] — is structurally capable of separating the two, and we did not test whether one does. The wider lesson is that identifying an undocumented step and measuring its effect is not the end of the exercise: the correction may not lie inside the space of choices the step appears to offer.

**Scale as a second hidden denominator.** Component-level analysis shows that detection improves with artifact size within every type, and that the types remain separated at every size band in which more than a handful of components exist. The consequence is not that small objects are hard, though they are. It is that the four types occupy almost non-overlapping ranges of object size, and that annotated area is concentrated in the largest components, so a pooled per-type sensitivity summarises a different part of the size range for each type. Comparing those numbers to each other is a category error of the same kind as comparing them across different coverage denominators.

The remedy is available and cheap. Reporting detected fraction as a function of component size, with the number of components in each band shown, makes the non-overlap visible and makes results comparable across benchmarks whose type composition differs — which any two benchmarks' will. Reporting the size distribution of the annotation alongside it costs one table and tells a reader which part of that function the headline number actually reflects. Neither requires any additional computation beyond connected-component labelling of the ground truth, which a benchmark could supply once and for all.

**Partition sensitivity.** Our results, and the original work's, rest on one partition. Surveying 200 random partitions of the same cohort shows that training composition is stable — no meaningful risk of a type being absent from training — but that the test side varies substantially, and that the partition in use is at the 7th percentile of effective sample size. Part of the statistical weakness reported here is a property of the draw rather than of the cohort. We regard this as the most useful single check we ran: it cost minutes of processor time, it corrected an over-strong claim we had already written, and it removed the case for a 30-GPU-hour experiment. It also illustrates the protocol's economics, since the cheap measurement determined that the expensive one was not worth making.

### 4.3 Generality and limits of the protocol

The protocol is demonstrated on one method and one benchmark. Three things bear on how far that demonstration reaches.

**The conditions are stated independently of the case.** Section 2.1.1 defines when the protocol is warranted in terms of properties a reader can check on their own resource: the number of independent units, the concentration of the target signal across them, whether the metric is a pooled ratio, and whether the partition is inherited. None of these refers to histopathology, to artifacts, or to diffusion models. Whether they are commonly satisfied is an empirical question we have not settled at scale — auditing a public archive against criteria of this kind is itself a piece of work [45] — but they are satisfied by every public whole-slide quality-control resource we are aware of, and the first three are satisfied by a large part of medical image segmentation, where exhaustive annotation is expensive [27] for the same reasons.

**Three of the four axes transfer without modification; the fourth requires judgement.** Resampling, seed repetition and partition surveying are mechanical and their cost does not grow with the domain. Identifying undocumented preprocessing does not reduce to a procedure: it requires knowing what a working implementation needs that the description does not supply, which is domain knowledge. What does transfer is the prior check — name the image property the gate operates on, and ask whether the phenomenon under study alters it. That question is answerable in any domain and, as the case study shows, can predict the outcome before any measurement is taken.

**The demonstration cannot establish frequency.** We have shown that on one benchmark satisfying the four conditions, three of four comparative claims did not survive. We have not shown what fraction of published claims on such benchmarks would fail, and no single case study could. The claim we make is conditional and modest: where the four conditions hold, these four checks are cheap enough to be routine, and in the one case examined they changed the conclusions materially. Establishing the frequency would require applying the protocol across several benchmarks and methods, which is the natural extension of this work.

### 4.4 Limitations

Our reconstruction is not the authors' code, and residual differences in unspecified details cannot be excluded; we mitigated this by training every documented ambiguity rather than one interpretation, but the space of undocumented choices is not enumerable. One such difference is known and large: our training set holds a twenty-ninth of the clean patches the source reports, because the parameters that determine the count are not given.

Each axis is instantiated more weakly than it could be. Seed variance rests on two runs of a single configuration, which yields a usable but imprecise estimate and assumes comparable variance in the other configurations. Partition variance was surveyed by composition rather than measured by retraining, a decision the composition data support but do not prove. And Axis 4 was instantiated by decomposing one restriction rather than by implementing a second: we localised the exclusion to the components of the composite and profiled the branches of its criterion on individual slides, and we argue on mechanistic grounds that a second cheap criterion would not differ, but we did not build one, and we did not test whether the learned tissue segmentation we recommend instead in fact retains blurred tissue. The spread we report is therefore a lower bound on the spread the axis is meant to capture.

The supervised baseline was unavailable for three frozen-section slides, all clean, which reduces power on the clean-slide false-positive rate.

And all of the above concerns fifty slides from a single archive. They are not homogeneous — they span four cancer types and, being TCGA material, a range of contributing sites and scanning systems [39], both of which are known to shift artifact content substantially. But they are one stain, one annotation source, and one retrospective research archive whose preparation and digitisation need not resemble routine clinical practice. Nothing here speaks to immunohistochemistry, to prospectively collected material, or to whether a second annotation team would have drawn the same boundaries — a source of variability we did not measure at all, and which on a benchmark this concentrated could plausibly rival the four we did.

## 5. Conclusions

Benchmarks with few independent units, concentrated ground truth, pooled ratio metrics and a single inherited partition are the normal condition of whole-slide quality control, and they support far weaker conclusions than the way results on them are ordinarily reported. We proposed a protocol that quantifies four sources of variability — test-set sampling, training stochasticity, partition composition, and undocumented preprocessing — and requires a claim to survive all four. Three of the four cost minutes of processor time and the fourth one training run.

Applied to an independent reconstruction of a published diffusion-based artifact detector, the protocol separates the findings. The method's central mechanism reproduces: the auxiliary contrastive term yields a small but real improvement that replicates across independently trained models, though it acts on pen marking rather than on the artifact types cited as its motivation. Its comparative claims do not: differences between design variants, and against a supervised baseline, lie inside the uncertainty of the evaluation, and the ordering against that baseline in fact reverses in our hands without either ordering being statistically supported.

The binding constraint is the benchmark rather than any method evaluated on it. Scored ground truth concentrated in four of 24 slides yields an effective sample size of six and a confidence interval of ±0.19 on any single pooled F1 — an interval that contains essentially every figure published on this resource. The partition in general use sits at the 7th percentile of effective sample size among partitions of the same cohort. And a tissue restriction that no published description specifies excludes 41% of out-of-focus annotation against 3% of air bubble, distorting comparisons between types.

That last finding is the one we would most want carried forward, because it does not depend on this benchmark, this method or this modality. The exclusion is not a poorly chosen threshold: cheap tissue criteria threshold the very image properties whose degradation defines out-of-focus tissue, so gate and artifact are confounded by construction and no adjustment within that family separates them. An undocumented step can leave a reconstruction with no correct choice among the choices it appears to offer — and whether it does can be established by asking what property the step measures, before any measurement is made.

None of this argues against the benchmark, which remains one of the few public resources of its kind, or against the method, whose mechanism we confirm. It argues that the four checks described here cost a small fraction of the training they accompany, and that between them they separate the effects that are real from the differences that are not.

## Figure captions

**Data**
50 H&E WSIs (TCGA), AIRAQC annotations, 4 evaluated artifact types

**Partition**
16 training / 24 test / 2 validation

♦ the partition is fixed
*our test set sits at the 7th percentile of effective sample size among 200 random partitions*

**Training**
2,185 clean patches extracted (63,000 specified) and 400 artifact;
LoRA on PixCell-1024, ± contrastive term

♦ contrastive term on $f_A$ or on the raw latent
*per-type shifts up to 0.080 with opposing signs; pooled F1 unchanged*

**Inference**
sliding window at 10x, $t^* = 800$, reconstruction-error heatmap

**Post-processing and metrics**
Gaussian, clip to $[v_{min}, v_{max}]$, Otsu, close-then-open; pixel-level F1

♦ $v_{min}$, $v_{max}$ described only as user-defined
*chosen by grid search on the test set: every reported F1 is a maximum over 25 points*

♦ no tissue mask specified
*out-of-focus sensitivity is 0.892 or 0.527, depending on how the mask is realised*

**Every claim must survive four independent sources of variability**

| | |
|---|---|
| **test-set sampling** | stratified paired bootstrap, 20,000 replicates |
| **training** | a second seed, identical data and thresholds |
| **partition** | composition survey over 200 random partitions |
| **preprocessing** | the undocumented choices above, decomposed exactly |

**Fig. 1** The reconstruction pipeline, the points at which the published description leaves a decision to the implementer, and the four axes of the reliability protocol applied to every claim. Diamonds mark choices the original account does not specify; the italic line beneath each gives the consequence measured in this study. Two further undocumented choices — the artifact-patch sampling scheme and the minimum annotation overlap defining an artifact patch — were also ablated and had no measurable effect; they are omitted here for space. The restriction of predictions to tissue, the choice with the largest measured consequence, is shown as a single node although it acts at two points in the pipeline: it determines which pixels are scored and, because the Otsu threshold is computed inside it, the decision threshold as well.

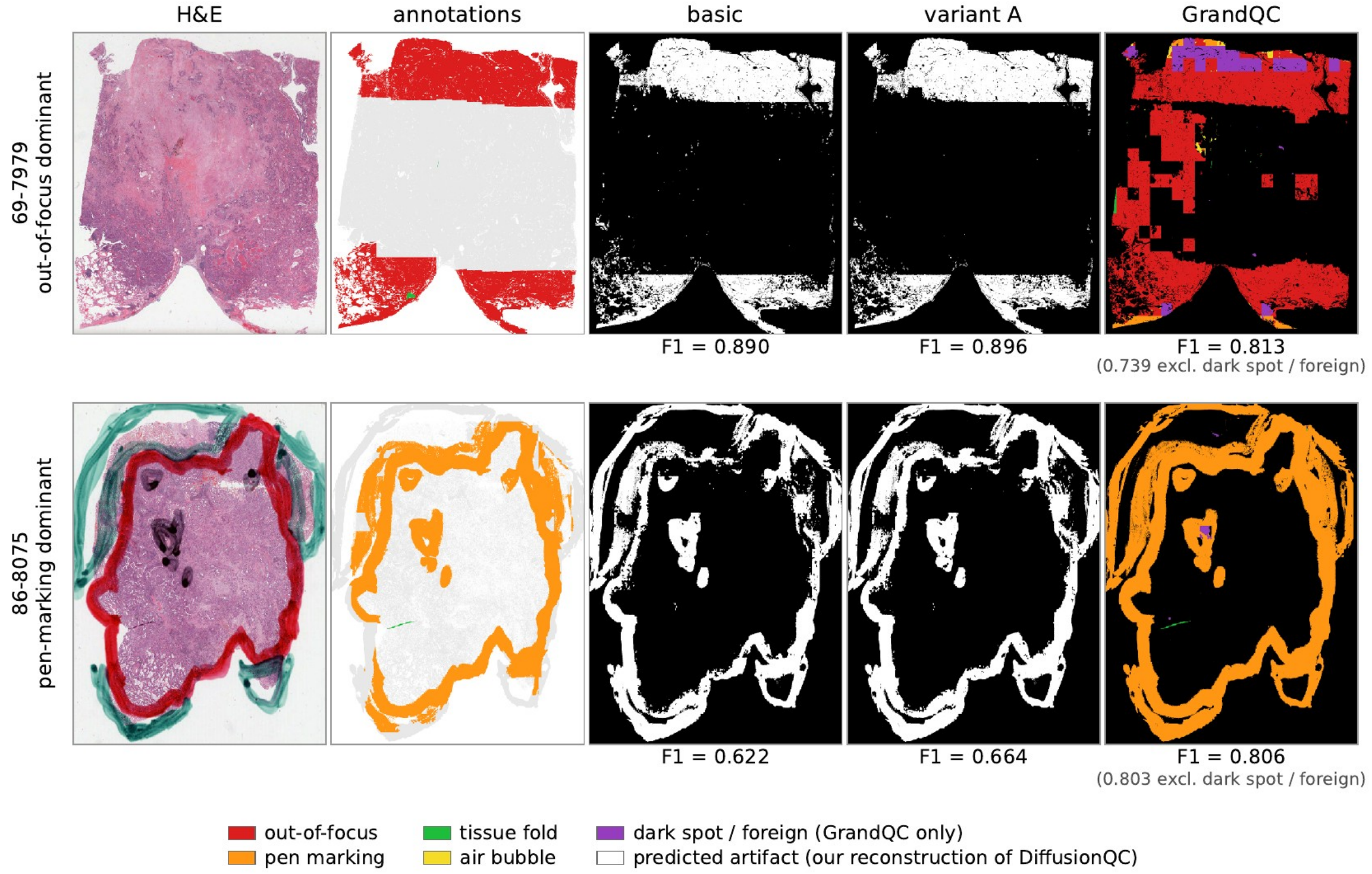

**Fig. 2** Qualitative comparison on the two test slides that appear in the original publication's figures, laid out to match it. Columns: H&E thumbnail; AIRAQC annotations by artifact type; the reconstructed basic model; the reconstructed variant A; GrandQC. Binary outputs are shown white-on-black because the reconstructed models emit a single artifact/not-artifact decision and do not distinguish types. GrandQC's classes are drawn in the AIRAQC colours so that the last column can be read against the second; its dark spot / foreign object class, which has no counterpart among the four evaluated types, is purple. Per-slide F1 is given beneath each prediction; for GrandQC the second value excludes the dark-spot channel. The two slides illustrate why that scoring choice is reported both ways: on TCGA-69-7979 the excluded channel covers visible area and costs 0.074, while on TCGA-86-8075 it is negligible and costs 0.003.

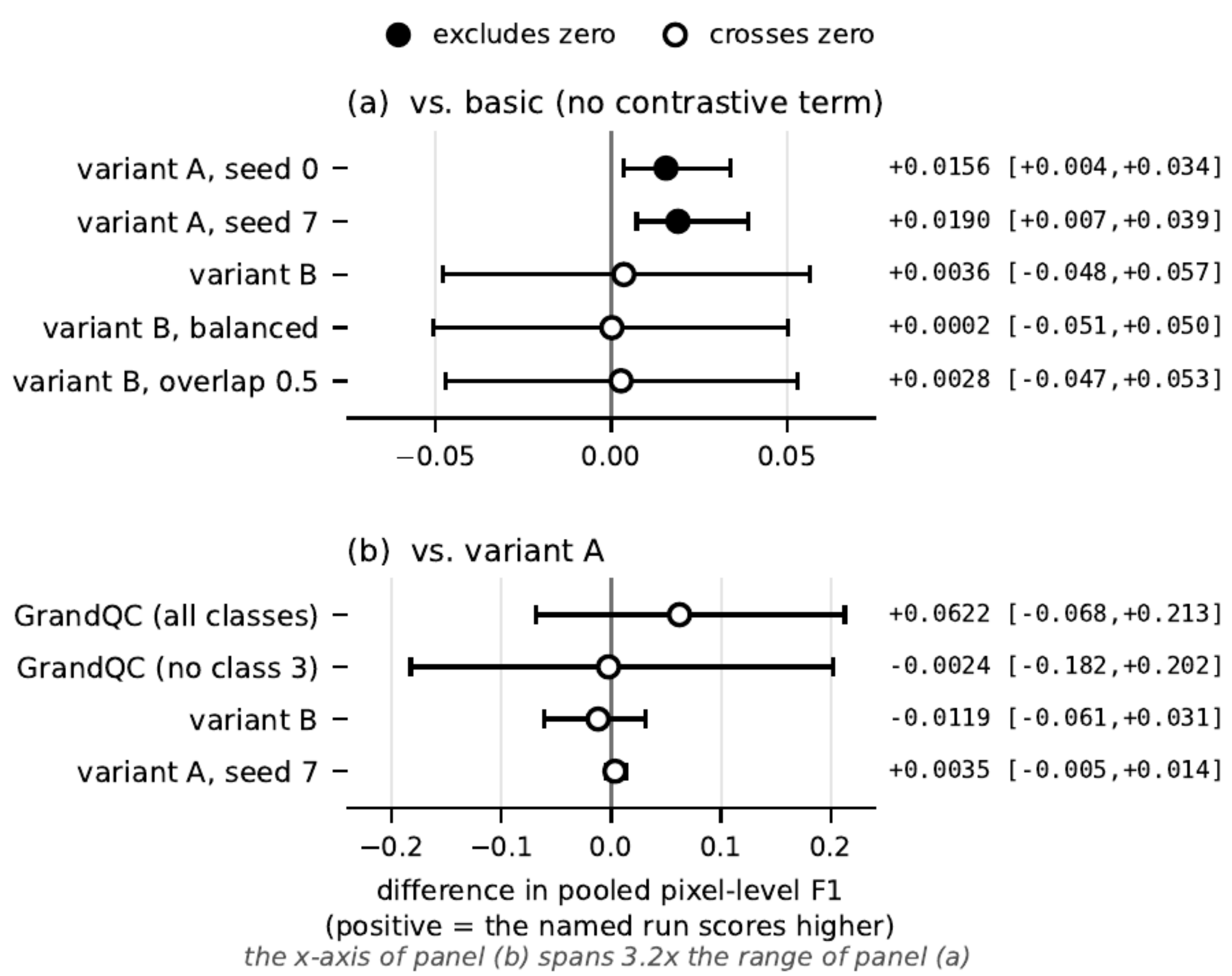

**Fig. 3** Paired differences in pooled pixel-level F1 with 95% percentile intervals from a stratified paired bootstrap (20,000 replicates, seed 0). Filled markers mark intervals excluding zero. **(a)** Every configuration against the basic model: only variant A separates from it, and it does so under both seeds. **(b)** Selected runs against variant A. Note that the x-axis of (b) spans 3.2 times the range of (a): pairing removes variance only to the extent that two runs agree slide by slide, so the interval against a closely related configuration is far narrower than the interval against a different method altogether. Two-sided paired permutation p-values (100,000 sign-flips) are, in panel (a) order, 0.031, 0.005, 0.884, 0.993, 0.907; in panel (b) order, 0.404, 0.983, 0.553, 0.463.

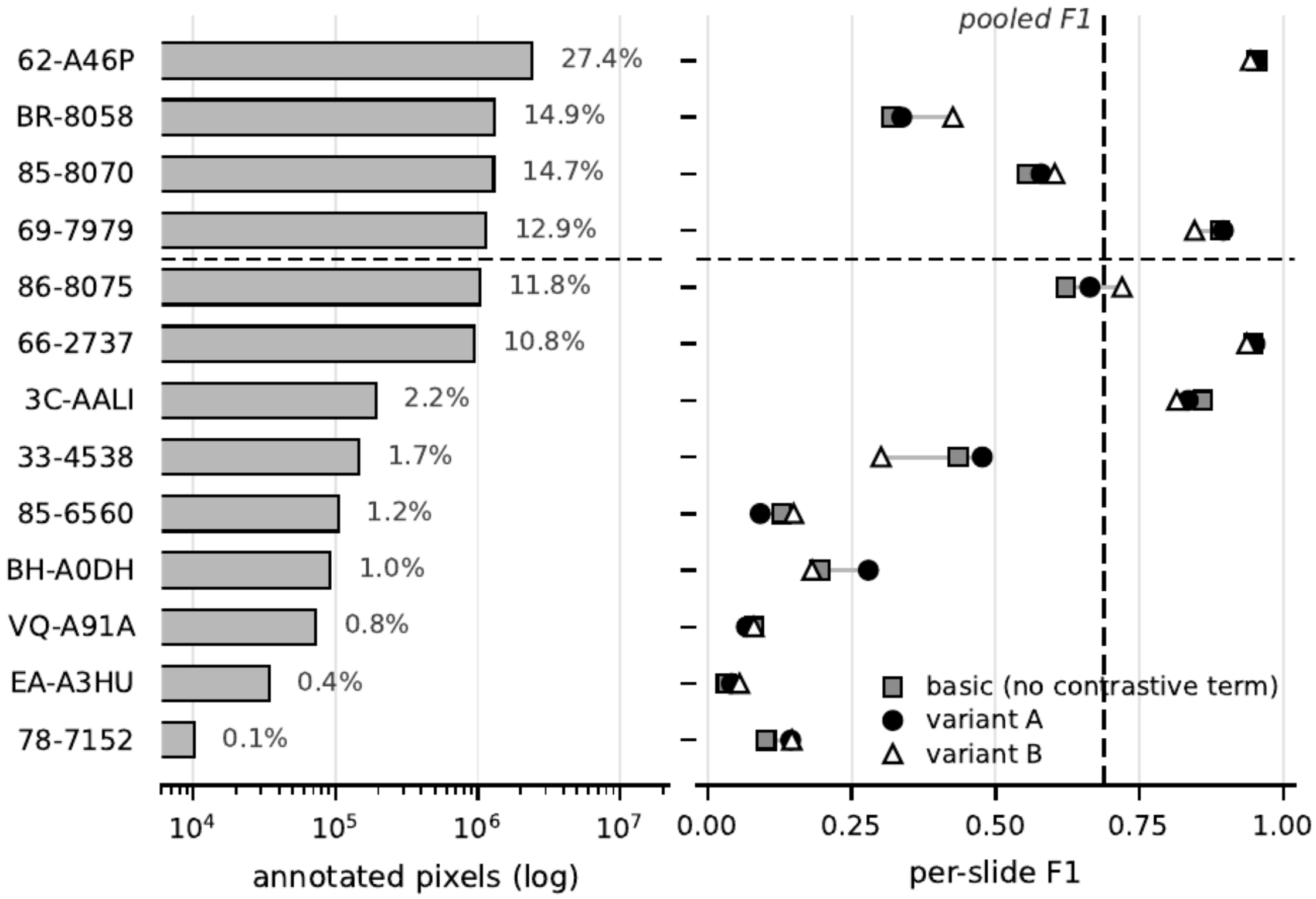

**Fig. 4** Why the intervals in Fig. 3 are as wide as they are, and why pairing narrows them. Thirteen artifact-bearing test slides, ordered by annotated area. **Left:** annotated pixels per slide on a log scale, with each slide's share of the total. **Right:** per-slide F1 for three configurations — basic, variant A, variant B — on the same rows, with the pooled value of variant A marked. Two things are visible at once: F1 ranges from 0.03 to 0.95 across slides and configurations, which is why resampling slides moves the pooled figure so much; and the three configurations sit almost on top of one another within every row, which is why the difference between them is estimated far more precisely than any of their levels. Almost no individual slide sits near the pooled value, which is a ratio of summed counts dominated by a few large slides rather than a typical slide's performance.

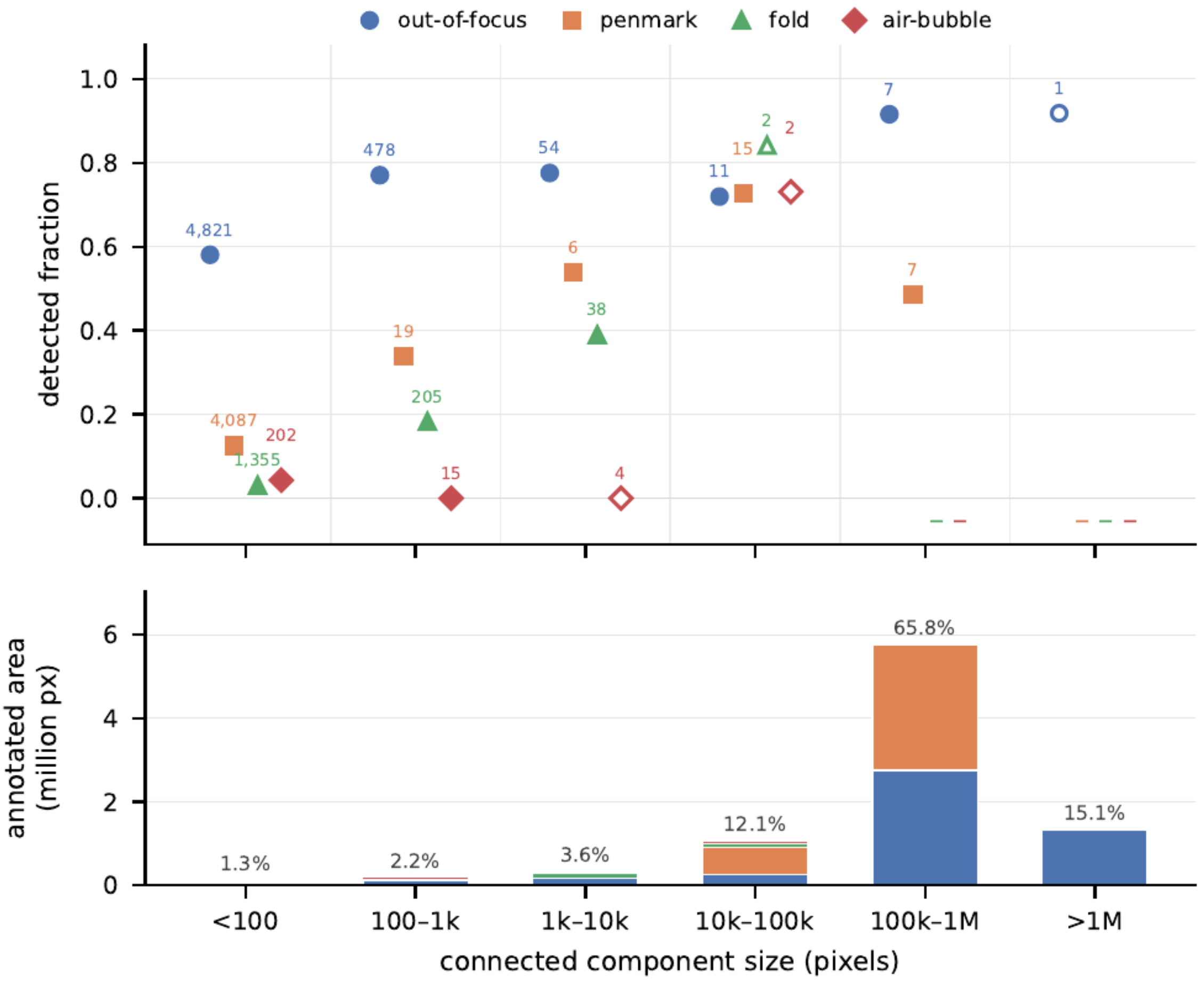

**Fig. 5.** Detection by component size and artifact type, variant A at t = 800. *Upper:* fraction of each annotated connected component recovered by the prediction, pooled within size bands. The number of components behind each estimate is printed beside it, and markers based on fewer than five components are drawn hollow. A dash on the baseline marks a type with no components in that band, which is distinct from a measured zero — air bubble scores 0.000 in the 100–1k and 1k–10k bands and is absent above $10^5$ pixels. *Lower:* annotated area per band, stacked by type. The types separate at every band containing a usable number of components, and they occupy different parts of the size range: folding and air bubble have no components above $10^5$ pixels, while 80.8% of annotated area lies above that threshold and belongs almost entirely to out-of-focus and pen marking. A pooled per-type sensitivity therefore summarises a different portion of this figure for each type.

## Declarations

**CRediT authorship contribution statement.** Konstantinos Moutselos: Conceptualization, Methodology, Software, Validation, Formal analysis, Investigation, Data curation, Visualization, Writing – original draft, Writing – review & editing. Ilias Maglogiannis: Conceptualization, Methodology, Supervision, Resources, Funding acquisition, Writing – review & editing.

**Declaration of competing interest.** The authors declare that they have no known competing financial interests or personal relationships that could have appeared to influence the work reported in this paper.

**Data availability.** All primary data are public: the whole-slide images from the TCGA data portal, the artifact annotations released by the AIRAQC authors, and the GrandQC quality-control masks for the TCGA cohort deposited at Zenodo (10.5281/zenodo.14041578, CC BY-NC-SA 4.0). The evaluation outputs generated in this study — per-slide confusion counts for every configuration reported, the bootstrap replicates underlying Figs. 3 and 4 and Tables 2 and 3, the partition survey, and the mask decomposition of Section 3.4 — are deposited together with the code at Zenodo (DOI: 10.5281/zenodo.22068659) [46]. Every interval reported in this paper can be recomputed from those counts without access to the whole-slide images. The deposit is released under the MIT licence for code and CC BY 4.0 for the derived data, as stated in the licence files it carries. The GrandQC masks themselves are not redistributed here; our deposit contains only counts derived from them, and the masks should be obtained from the record cited above under their own licence.

**Code availability.** The method was reconstructed from its published description; no reference implementation was available. The reconstruction, the evaluation pipeline, the bootstrap and permutation code, the partition survey and the diagnostics of Section 3.4 are available in the same Zenodo record (DOI: 10.5281/zenodo.22068659) [46], together with the exact post-processing parameters for every result reported, so that each evaluation can be replayed deterministically. The archive carries an explicit notice that it is not a reference implementation of the method it reconstructs.

**Funding.** This work was carried out within Project PRODIGY, Greek–Chinese Bilateral Research and Technology Cooperation, Grant No. ΔΣΕΚΡ01-0021403, Programme Competitiveness 2021–2027, Ministry of Economy and Finance, Greece, ERDF, co-funded by the European Union. Computing resources were provided by the University of Piraeus.

**Ethics.** This study used publicly available, de-identified whole-slide images from The Cancer Genome Atlas and publicly released annotations derived from them. No new human data were collected and no ethics approval was required.

**Acknowledgements.** The authors thank the AIRAQC and GrandQC groups for releasing their annotations and masks publicly, without which this study would not have been possible.

**Declaration of generative AI and AI-assisted technologies in the manuscript preparation process.** During the preparation of this work the authors used a large language model in order to draft and revise the manuscript text. After using this tool, the authors reviewed and edited the content as needed and take full responsibility for the content of the published article.